\documentclass[journal]{IEEEtran}
\usepackage{cite}
\usepackage{amssymb}
\usepackage{amsmath}
\usepackage{amsfonts}
\usepackage{color}
\usepackage{algorithmic}
\usepackage{algorithm}
\usepackage{amsthm}
\usepackage{graphicx}
\usepackage{epstopdf}
\usepackage{subfig}
\usepackage{booktabs}
\usepackage{lipsum}
\usepackage{cuted}
\usepackage{textcomp}
\usepackage{verbatim}
\usepackage{array}  
\usepackage{float}
\usepackage{bm}
\usepackage{xcolor}
\usepackage{subfig}
\usepackage{multicol}
\usepackage{hyperref}

\ifCLASSINFOpdf

\else

\fi
\begin{document}
	\setlength{\abovedisplayskip}{2pt}
	\setlength{\belowdisplayskip}{2pt}
       	\title{AIGC-assisted Federated Learning for Edge Intelligence: Architecture Design, Research Challenges and Future Directions}
        \author{Xianke Qiang, Zheng Chang,~\IEEEmembership{Senior~Member,~IEEE}, Ying-Chang Liang,~\IEEEmembership{Fellow,~IEEE}
	   \thanks{X. Qiang and Z. Chang are with School of Computer Science and Engineering, University of Electronic Science and Technology of China, Chengdu 611731, China. Y-C. Liang is with Center for Intelligent Networking and Communications (CINC), University of Electronic Science and Technology of China, 611731 Chengdu, China.}}
        \maketitle
\begin{abstract}
     Federated learning (FL) can fully leverage large-scale terminal data while ensuring privacy and security, and is considered as a distributed alternative for the centralized machine learning. However, the issue of data heterogeneity poses limitations on FL's performance. To address this challenge, artificial intelligence-generated content (AIGC) which is an innovative data synthesis technique emerges as one potential solution. In this article, we first provide an overview of the system architecture, performance metrics, and challenges associated with AIGC-assistant FL system design. We then propose the Generative federated learning (GenFL) architecture and present its workflow, including the design of aggregation and weight policy. Finally, using the CIFAR10 and CIFAR100 datasets, we employ diffusion models to generate dataset and improve FL performance. Experiments conducted under various non-independent and identically distributed (non-IID) data distributions demonstrate the effectiveness of GenFL on overcoming the bottlenecks in FL caused by data heterogeneity. Open research directions in the research of AIGC-assisted FL are also discussed.
\end{abstract}
	
	
\section{Introduction}
	With the advent of 5G, the Internet of Things (IoT), Artificial Intelligence (AI), and many other emerging technologies, vast amount of data are generated at the network edge, providing a solid foundation for the evolution of many intelligent services\cite{10528244}. These services now offer multi-modal functionalities, integrating various forms of content such as text, audio, images, and video, transcending traditional limitations. However, rising concerns about privacy and limitations in network bandwidth have constrained the traditional approach of collecting data from distributed clients for centralized model training. These problems make the centralized data processing model become challenging, highlighting the need to explore more decentralized and privacy-preserving data processing methods.\par
    
    Federated Learning (FL) is a transformative paradigm that enables multiple clients to collaboratively train a shared global model while keeping their local data decentralized. This approach enhances data privacy and security, addressing the challenges associated with data sharing across diverse environments \cite{li2024filling}. However, FL encounters significant challenges due to the heterogeneity of data and resources among clients.\par
    
\begin{figure}[tbp]
        \centering
        \includegraphics[scale=0.68]{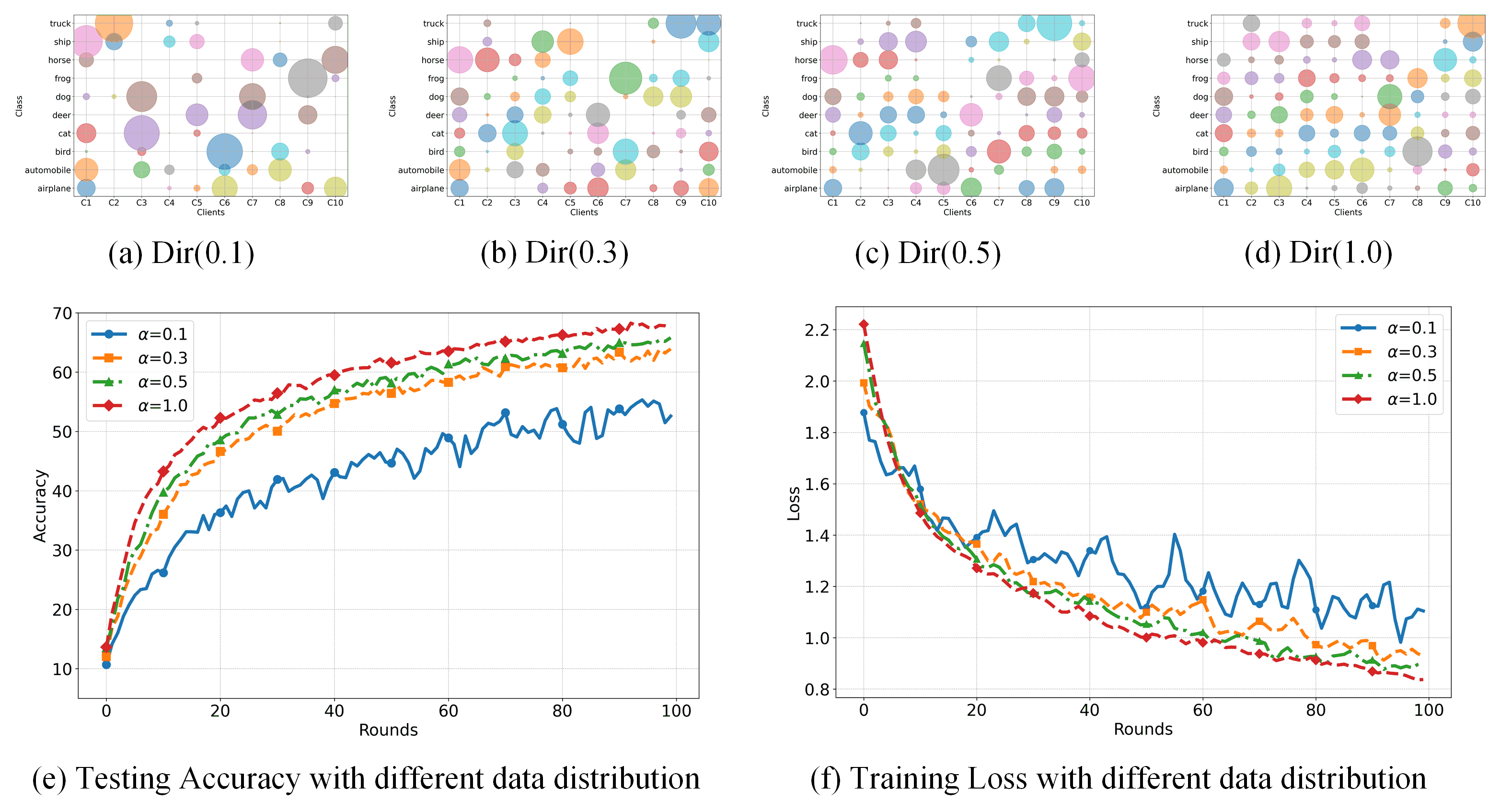}
        \caption{The impacts of data distribution on the training performance.}
        \label{fig: nonIID} 
    \end{figure}

    A major challenge in FL is the presence of non-independent and identically distributed (non-IID) data, which can hinder the convergence of the global model and lead to suboptimal performance. To investigate this issue, we apply different degrees of Dirichlet distribution to the CIFAR-10 dataset, introducing varying levels of non-IID data\cite{wang2020federated}. As shown in the top subplot of Fig. \ref{fig: nonIID}, a well-distributed Dir(1.0) results in better convergence accuracy and rate compared to Dir(0.1), highlighting the critical need to address non-IID data challenges.\par

    To tackle issues related to non-IID data distribution and data scarcity in FL, the rapid development of artificial intelligence-generated content (AIGC) services, such as Stable Diffusion (SD), DALL-E2, and Imagen, offer a promising solution \cite{10445209}. These generative AI techniques enable clients to quickly produce high-quality synthetic data (e.g., images and videos), effectively supplementing local datasets and mitigating the inherent limitations of FL. Moreover, AIGC can facilitate multi-modal FL by generating data across different modalities, such as images, text, and audio. This capability significantly enhances the model's performance in processing and integrating cross-modal information, making FL more versatile for applications like healthcare, autonomous driving, and smart cities. Additionally, AIGC supports few-shot learning and transfer learning in FL. By generating representative synthetic data, it helps address the challenge of limited data availability on client clients, thereby improving the model's generalization and adaptability across various tasks and environments.\par
    \begin{figure}[tbp]
        \centering
        \includegraphics[scale=0.55]{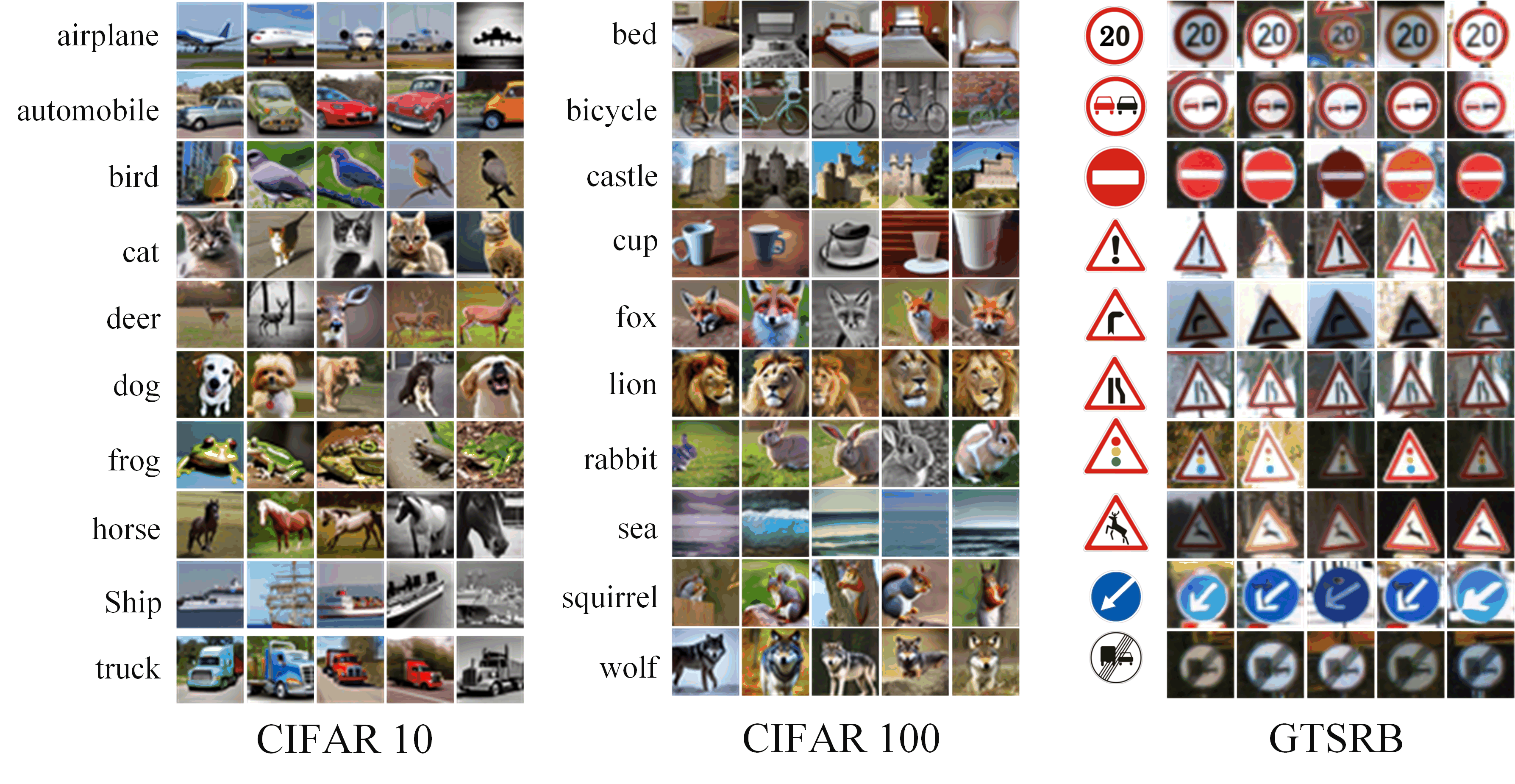}
        \caption{Examples of generated images.}
        \label{fig: generatedimages} 
    \end{figure}
    
    Recent studies \cite{10433716, jeong2018communication, 10557146, li2024filling, 10630904} have proposed data augmentation techniques that generate data to improve clients' local datasets. These approaches allow clients to use generative AI to produce images that augment their local data, thus mitigating the negative effects of non-IID data distributions. However, the augmented training data introduces extra computation costs for clients. There are still many challenges to address within the AIGC-assisted FL system: 
    \begin{itemize}
        \item \textit{First, how to design an efficient AIGC-assisted FL system?} Not all clients have the capability or willingness to generate images, as they may prioritize conserving energy for their core functions. Therefore, it is essential to design an AIGC architecture that supports FL while minimizing client burden, addressing non-IID challenges, and ensuring privacy protection. Despite recent advancements, several challenges remain in the AIGC-assisted FL system.
        \item \textit{Second, how to design an effective model augmentation strategy?} As shown in Fig. \ref{fig: generatedimages}, the images generated using diffusion model for the CIFAR10, CIFAR100, and GTSRB datasets exhibit impressive quality. Developing an effective strategy to integrate these generated images into FL models is crucial for improving their performance.
        \item \textit{Third, how to design an efficient computation and communication resource allocation strategy in AIGC-assistant FL?} Given the limited computational resources and constrained wireless bandwidth, it is crucial to develop a resource allocation strategy that maximizes the effectiveness of the AIGC-assistant FL system.
    \end{itemize}
    \par
    Keeping in mind these critical problems, this paper introduces Generative Federated Learning (GenFL) as an innovative solution to address the limitations of AIGC-assisted FL. GenFL is specifically crafted to tackle the complex issues of data heterogeneity, computational constraints, and resource optimization in federated systems. By integrating generative models with FL, we aim to not only enhance data diversity but also establish a more robust framework for effectively handling non-IID data distributions.\par
    To the best of our knowledge, this is the first magazine study to provide an overview of AIGC-assisted FL. The rest of this paper is organized as follows: we begin by presenting the architecture of the AIGC-assisted FL system, followed by an exploration of key performance metrics and associated challenges. We then demonstrate the effectiveness of our approach, and conclude with a discussion of potential future research directions.

\section{AIGC-assistant FL: Architecture, Metrics and Challenges}
    In this section, we first introduce AIGC-assistant FL system architecture and workflow. Then we introduce performance metrics from the aspects of data, model, system levels. Finally, we analyze the facing challenges in the AIGC-assistant FL system.
    \begin{figure*}[t]
        \centering
        \includegraphics[scale=0.95]{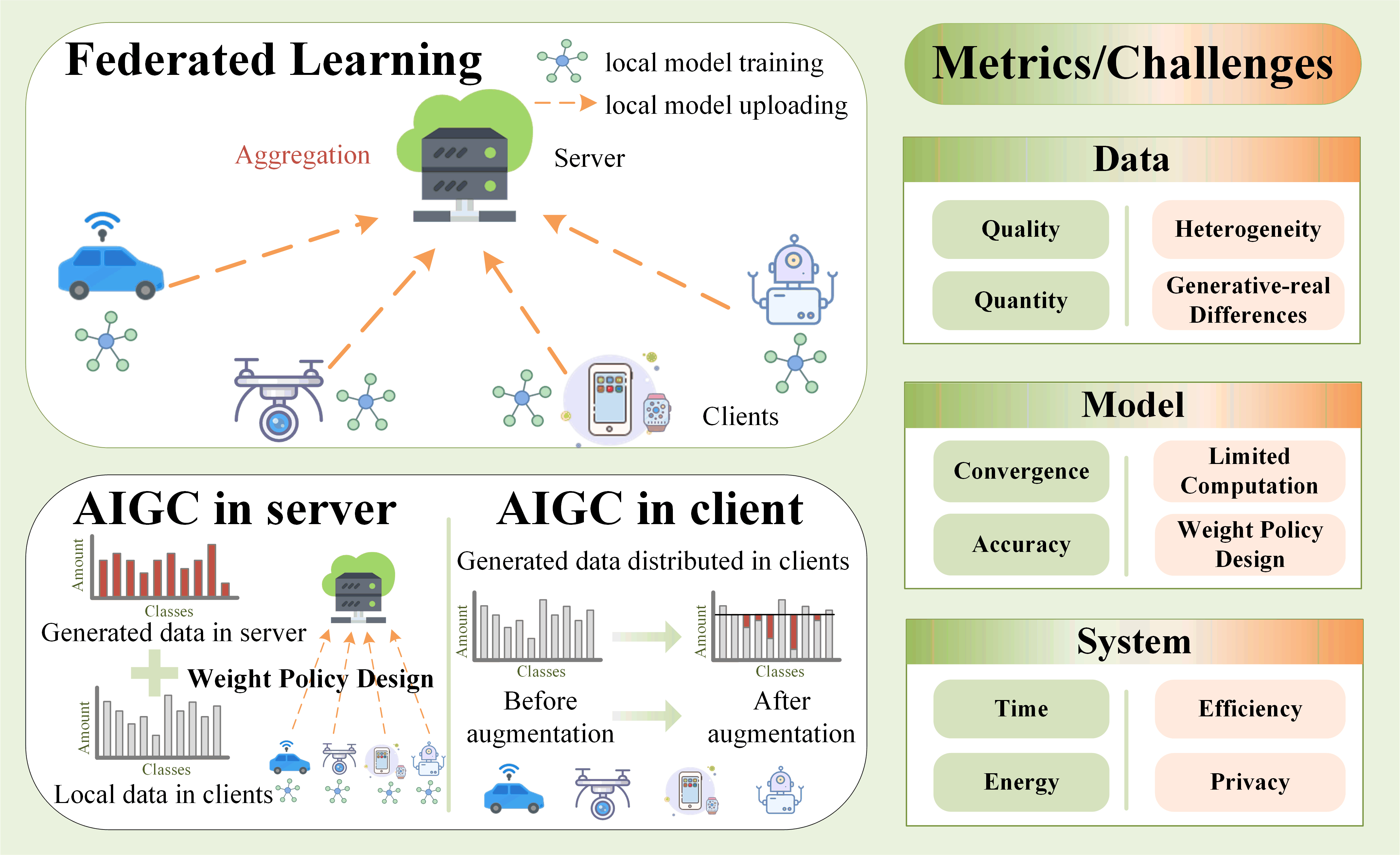}
        \caption{AIGC-assistant FL: architecture, metrics and challenges.}
        \label{fig: architecture} 
    \end{figure*}
    \subsection{System Architecture}
        We consider a general FL system comprising a server and a set of clients, as illustrated in Fig. \ref{fig: architecture}. Each client holds its own original local dataset. In the FL system, every client relies solely on its local dataset for model training and updates. The local datasets across all clients are distributed in a non-IID manner. AIGC-assisted FL can be categorized into two types: server-side AIGC and client-side AIGC. \par
        In the server-side AIGC scenario, all images generated by AIGC are stored on the server, while the clients retain their heterogeneous local datasets. This architecture allows the server to leverage its robust computational capabilities and rich data resources to generate and store diverse images during the aggregation process. In this context, to fully utilize the server-generated dataset to assist traditional FL, it is essential to design appropriate aggregation strategies and weight policies. This ensures effective integration of server-generated data with local client data during model training, ultimately enhancing the overall performance and accuracy of the global model.\par
        Client-side AIGC primarily involves two scenarios: one where images are generated directly on the client, and the other where images generated on the server are disseminated to the clients. In these cases, the client's dataset consists of the local dataset and the generated dataset. By effectively optimizing and integrating these two types of data, we can alleviate the issue of data heterogeneity to a certain extent, improving the representativeness and diversity of the training data. Moreover, this approach can enhance the client's learning capabilities, enabling better adaptation to changing environments and task requirements, thereby increasing the overall efficiency and robustness of federated learning.

    \subsection{Performance Metrics} 
    To evaluate the performance of the AIGC-assistant FL system, we consider the following metrics. At the data layer, we assess the quality and quantity of data to evaluate the heterogeneity of non-IID distributions. From the FL model perspective, we measure performance through convergence and accuracy. At the system level, we evaluate the system's cost in terms of time and energy, as the AIGC-assistant FL system combines both communication and computation within a distributed network framework.

        \subsubsection{Quality $\&$ Quantity}
        In FL systems, data is widely distributed across different clients. When addressing data heterogeneity, two key factors are considered: data quantity and quality. First, regarding data quantity, each client stores different categories of data, with the number of samples within each category varying. Assuming uniform data quality, the Earth Mover’s Distance (EMD) between the class distribution on each client and the overall population distribution can be used to quantify this heterogeneity \cite{zhao2018federated}. Besides the quantity of data, the quality of data is equally important. For large local datasets, quality differences mainly arise from variations in data collection clients and angles, resulting in some data being of lower quality or invalid. For datasets generated by AIGC, quality discrepancies often arise from ambiguities in interpretation. For instance, "apple" might refer to either a fruit or an iPhone, and "beetle" could be interpreted as either an animal or a "Beetle" car. Moreover, data quality differences are influenced by the differences between real and generated data. Local learning error \cite{li2024filling} and the upper bound of the gradient difference between the local and global loss functions \cite{10557146} are commonly used to assess data loss, which can be used to estimate data quality.
       
        \subsubsection{Convergence $\&$ Accuracy}
        Our goal is to train an optimal global FL model by aggregating the local models trained on various clients. However, the heterogeneity in local data leads to updates from each client becoming nearly orthogonal \cite{9107235}, which weakens the effectiveness of global updates and slows down model convergence. The convergence of the model is a key factor, as it reflects the stability and efficiency of the training process, ensuring the model reaches an optimal state. Training accuracy is an important indicator of the model’s learning capacity, reflecting its performance on the training data. On the other hand, testing accuracy serves as a critical metric for evaluating the model’s generalization ability, showing how well it performs on unseen data.

        \subsubsection{Time $\&$ Energy}
        In FL, it is crucial not only to focus on model convergence and accuracy but also to consider the overall system overhead. The system operates by connecting various clients to a central server through a wireless network, which introduces delays during local training and increases energy consumption. Specifically, local clients frequently communicate with the server during model training, contributing to network congestion and potential data transmission delays, which can impact the system's overall responsiveness. Additionally, continuous data exchanges can lead to inefficient use of computational resources, further reducing training efficiency. This inefficiency not only extends training times but may also hinder the model’s adaptability in dynamically changing data environments. Given these challenges, latency and energy consumption are critical factors that affect the performance of federated learning systems. Consequently, optimizing communication mechanisms is essential to minimize delays and energy usage, thereby improving the system’s efficiency and responsiveness.
        
    \subsection{Challenges} 
        In this subsection, we focus on the challenges faced by AIGC-assistant FL system from different layers. 
        \subsubsection{Data Layer} 
        The AIGC-assisted FL system faces significant challenges, primarily due to the heterogeneity of local data and the difference between generated data and real data. AI technologies rely on vast amounts of training data, and in FL, extensive local datasets are distributed across clients. However, the heterogeneity among these local datasets creates bottlenecks in the accuracy of FL models. Furthermore, not all local data is useful or of high quality. The differences in data distribution and quality pose challenges for the aggregation of the global model. Specifically, the disparities in local data may lead to client updates that are nearly orthogonal, weakening the effectiveness of global updates and slowing down the model's convergence\cite{charles2021large}. Moreover, inherent differences exist between images generated by AIGC and real images. While the generated images may appear visually realistic, they often exhibit significant feature discrepancies compared to real-world images. These differences can affect model performance, as the generated data may not fully capture the underlying characteristics found in real-world data.

        \subsubsection{Model Layer}
            In AIGC-assisted FL systems, two primary challenges arise at the model level: limited resource and weight policy design. \par
            First, limited computational resources pose a critical issue. In FL, while deeper local models can learn local datasets more effectively, training larger and deeper AI models on clients still presents significant challenges. Despite numerous studies attempting to alleviate the burden of model training through techniques such as quantization, pruning, and partitioning, these challenges remain unresolved. In AIGC-assisted FL, a common approach is to use AIGC models at the clients to generate images in order to mitigate the heterogeneity of local datasets. However, this practice is not cost-effective, as it consumes substantial energy, and the primary task of clients is not data generation. Therefore, excessively sacrificing the energy of clients to support data generation is inadvisable.\par
            Second, the design of weight policies presents considerable challenges. In FL, effectively aggregating model updates from different clients is a critical issue. The heterogeneity of client data can lead to differences in model updates, resulting in a decline in the performance of the global model. Therefore, designing an appropriate weight policy that accurately reflects the contribution of each client is essential for enhancing the accuracy and robustness of the global model. Optimizing this strategy involves not only the selection of algorithms but also a careful consideration of the resource status and data quality of each client.
    
        \subsubsection{System Layer}
            At the system level, AIGC-assisted FL system encounters a range of challenges during model training. \par
            First, limited computational resources pose a critical issue. The computational capabilities of clients are often constrained, making it extremely difficult to train complex AI models. Additionally, the heterogeneity of computing platforms presents significant challenges, as differences in computational capabilities among various client types lead to instability in system operations. Utilizing AIGC to generate images at the end to alleviate the heterogeneity of local datasets is not economical, as it consumes substantial energy, and the primary task of clients is not data generation. Therefore, excessively sacrificing the energy of clients to support data generation is unnecessary.\par
            Second, unstable wireless channels represent a significant threat to system performance. In distributed learning, high-performance wireless networks are crucial for accelerating the implementation of intelligent services, as intermediate parameters of model training must be transmitted over wireless networks. In FL, if data generated by AIGC on the server is then transmitted to the clients, this approach is also not economical, as it consumes a considerable amount of communication resources. However, due to the inherent instability of wireless networks, the large number of participating clients, and the mobility of some clients, not all clients can access sufficient communication bandwidth, and interruptions in communication may occur, pausing the training process. Therefore, there is a need to design resource management strategies to optimize the application of AIGC-assisted FL systems on clients.

\begin{figure*}[t]
    \centering
    \includegraphics[scale=0.95]{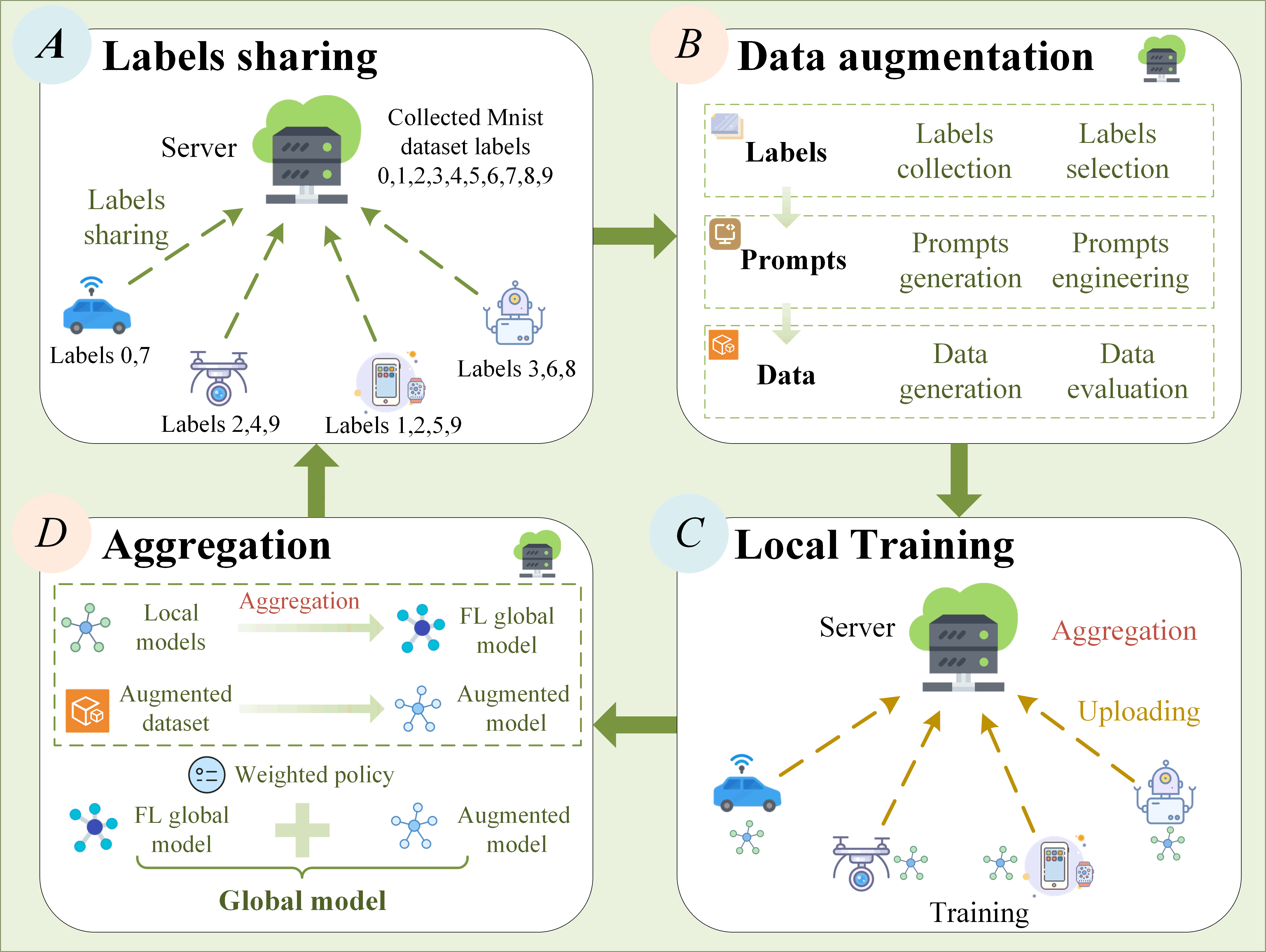}
    \caption{Workflow of proposed AIGC-assistant FL architecture.}
    \label{fig: workflow} 
\end{figure*}

\section{GenFL: AIGC in server}
     We consider a general FL system that includes one server and a set of clients $\mathcal{N}=\{1,2,\dots, n\}$ shown in Fig. \ref{fig: workflow}. Each client, denoted as $n$, holds an original local dataset $\mathcal{D}_n^{loc} = \{\mathcal{X}_n, \mathcal{Y}_n\}$, where $\mathcal{X}_n = \{x_{n}^{1}, x_{n}^2, \dots, x_{n}^{\left|\mathcal{D}_n\right|}\}$ is the training data, $\mathcal{Y}_n = \{y_{n}^1, y_{n}^2, \dots,y_{n}^{\left|\mathcal{D}_n\right|}\}$ represents the corresponding labels, and $|\mathcal{D}_n^{loc}|$ represents the number of training data samples of client $n$. We will use generative model to generate additional dataset in server, $\mathcal{D}^{gen}$. In FL system, client $n$ is solely relies on its local dataset $\mathcal{D}_n^{loc}$ for local model training and update. All clients' local dataset is distributed non-IID.
    \subsection{Labels Sharing}
    All clients can share the labels they held. The purpose of this step is to provide the server with a comprehensive overview of the clients' current states, which serves as a foundation for client selection and resource allocation. By sharing these labels, clients enable the server to form an effective and adaptive communication and computation strategy based on these information.
    \subsection{Data Augmentation}
    Once the server has gathered the labels and relevant information from the clients, it selects the labels to be generated. These selected labels are then used to create corresponding prompts. Finally, AIGC techniques, such as Stable Diffusion, are applied to generate a large number of images based on these prompts.
    \subsection{Local Training}
    While the server generates images, clients download the initial global model from the server and begin training their local models using their local datasets $\mathcal{D}_n^{loc}$ respectively. Then, clients upload updated local model to server for aggregation.
    \subsection{Aggregation and Weighted Policy}
    The server trains an augmented model $\bm{\omega}_a$ using $\mathcal{D}^{gen}$ dataset. Additionally, the server receives the updated local models $\bm{\omega}_n$ from the clients. The weighted policy for round $t+1$ is defined as 
        \begin{align}        
            \bm{\omega}^{t+1} = \kappa_1 \sum_{\forall n \in \mathcal{N}} \rho_n \bm{\omega}_n^t + \kappa_2 \bm{\omega}_a^t,  
            \label{modelaugmentation}
        \end{align}
    where $\kappa_1$ represents the proportion of the FL model, $\kappa_2$ corresponds to the proportion of the augmented model, and $\rho_n = \frac{\left|\mathcal{D}_n^{loc}\right|}{\sum_{\forall n \in \mathcal{N}} \left|\mathcal{D}_n^{loc}\right|}$. 

    \begin{figure*}[ht]
        \centering
        \subfloat[Dir($\alpha=0.1$)]{\includegraphics[width=0.33\textwidth]{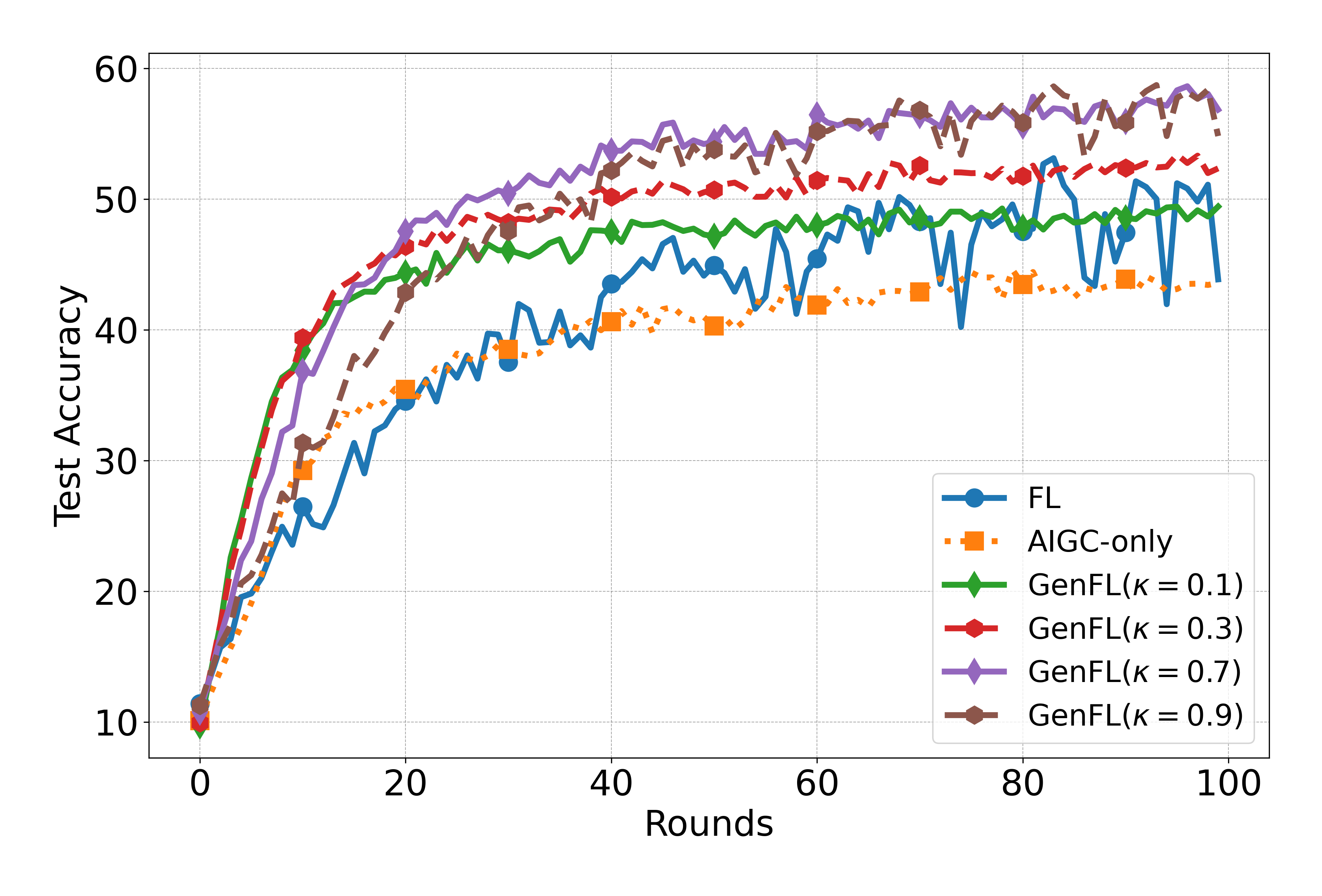}}
        \hfill
        \subfloat[Dir($\alpha=0.3$)]{\includegraphics[width=0.33\textwidth]{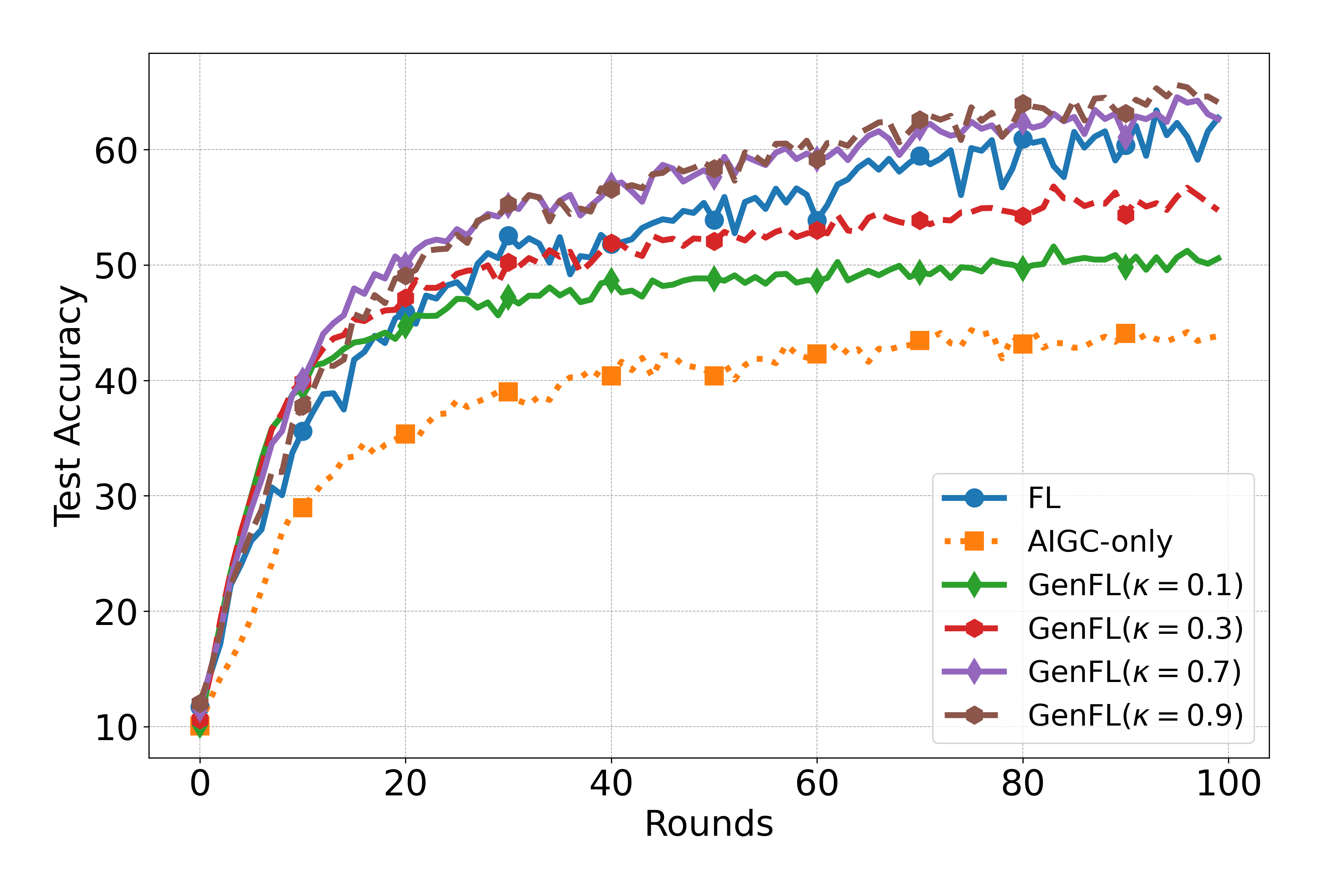}}
        \hfill
        \subfloat[Dir($\alpha=1.0$)]{\includegraphics[width=0.33\textwidth]{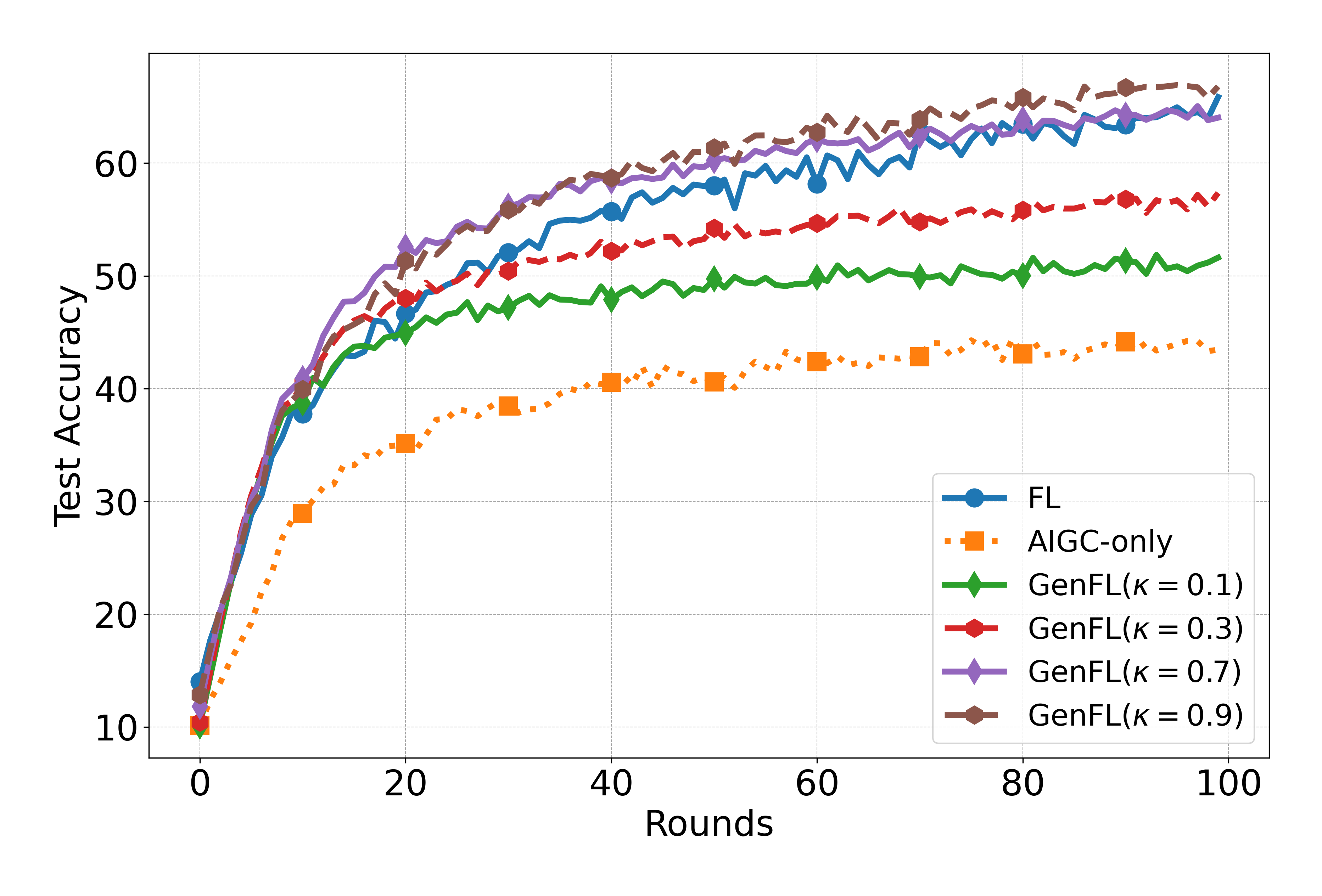}}
        \caption{Accuracy on CIFAR-10 with different Dirichlet distribution.}
        \label{fig:cifar10-acc}
    \end{figure*}
    \begin{figure*}[ht]
        \centering
        \subfloat[Dir($\alpha=0.1$)]{\includegraphics[width=0.33\textwidth]{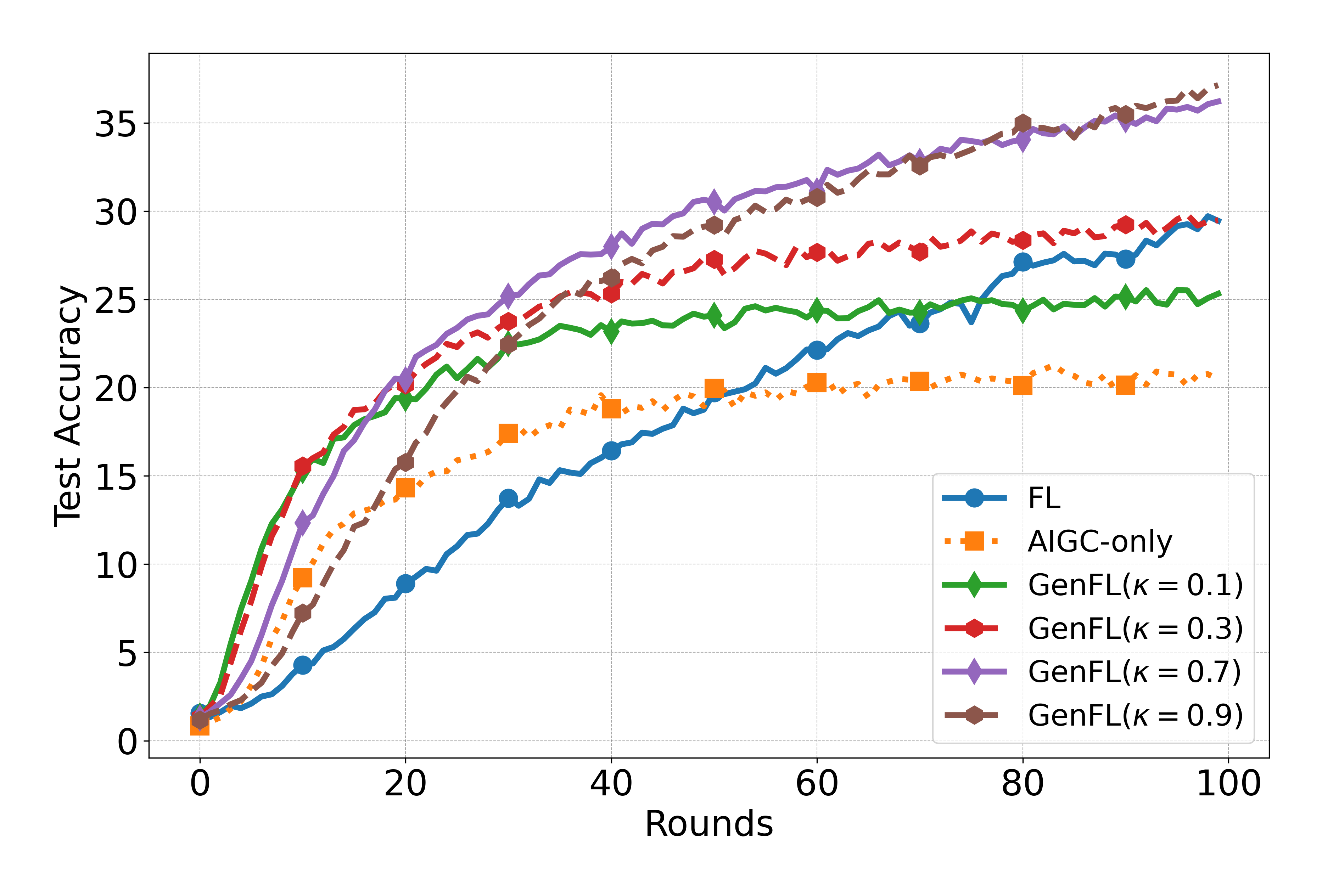}}
        \hfill
        \subfloat[Dir($\alpha=0.3$)]{\includegraphics[width=0.33\textwidth]{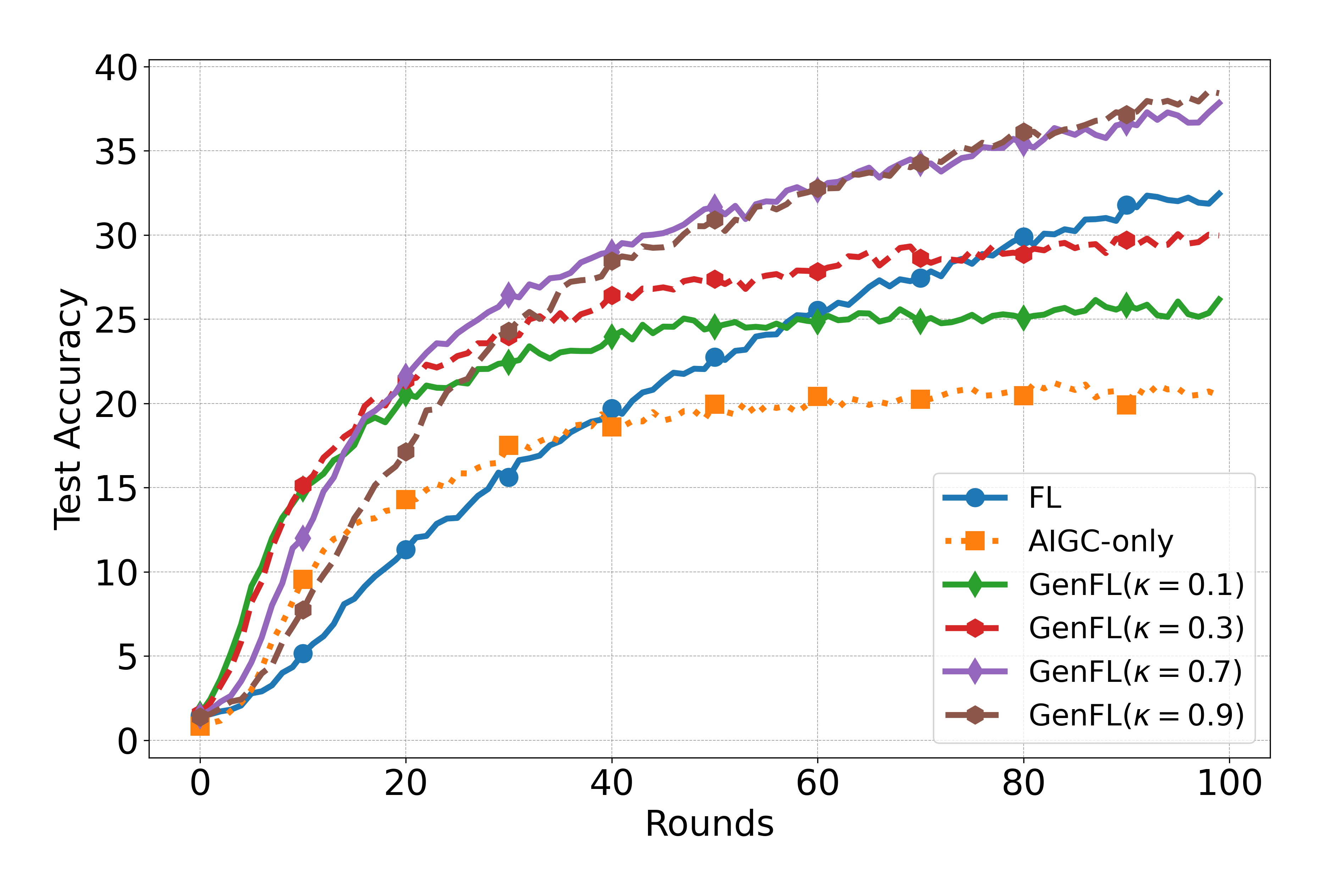}}
        \hfill
        \subfloat[Dir($\alpha=1.0$)]{\includegraphics[width=0.33\textwidth]{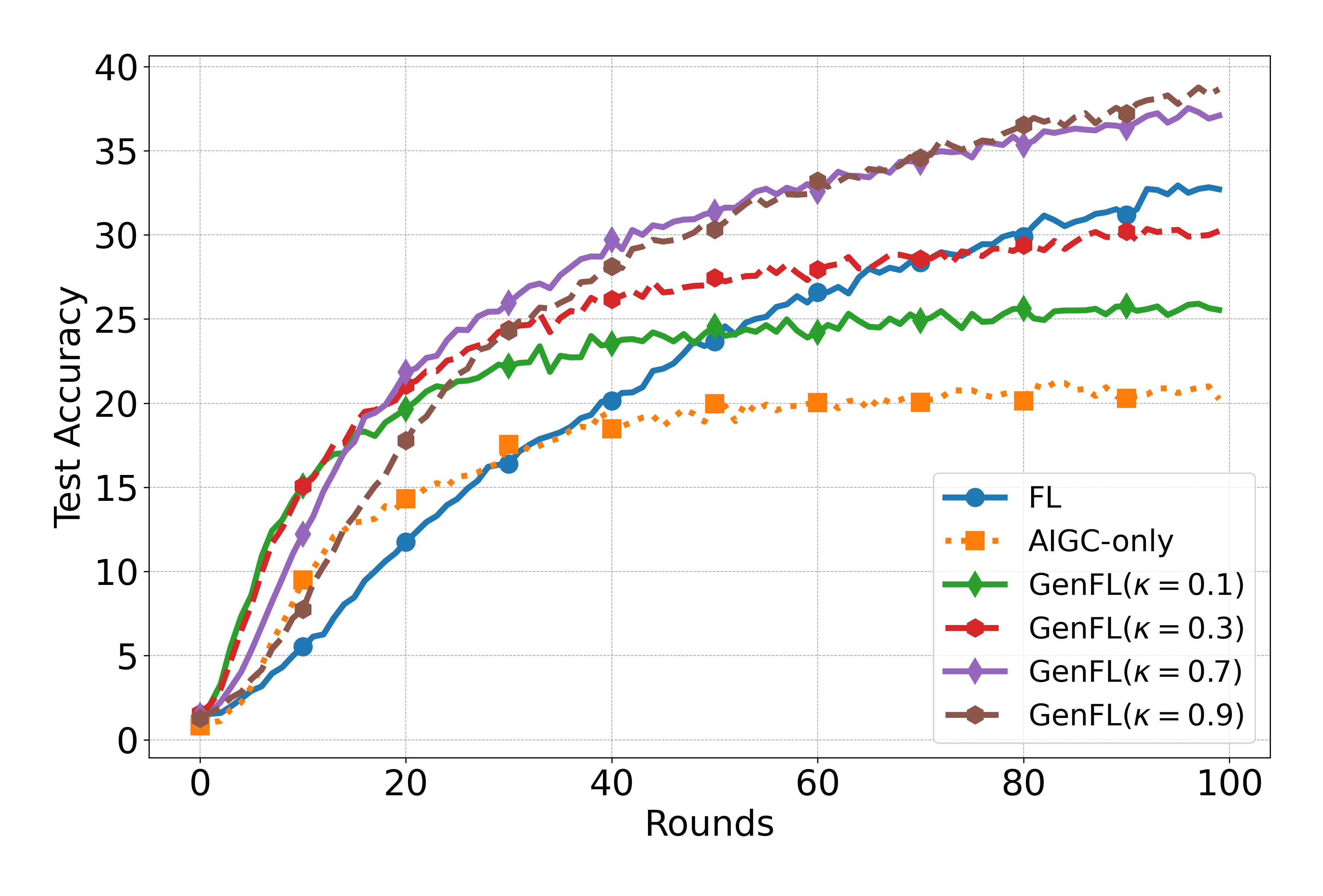}}
        \caption{Accuracy on CIFAR-100 with different Dirichlet distribution.}
        \label{fig:cifar100-acc}
    \end{figure*} 
  
\section{Performance Evaluation}
    In this section, we evaluate the performance of the proposed weighted policy in the GenFL architecture under different Dirichlet data distribution using the CIFAR-10 and CIFAR-100 datasets.
    \subsubsection{Experiment Setting}
        In this subsection, we will introduce the details of our experimental setup, including the use of open datasets, the FL configuration, the generation setting, and the prompt templates. \par
\begin{itemize}
    \item{Open datasets:} Our simulation leverages two distinct image classification datasets: (1) the CIFAR-10 dataset \cite{krizhevsky2009learning}, containing colored images categorized into 10 classes such as "Airplane" and "Automobile". CIFAR-10 dataset comprises a training set with 50,000 samples and a testing set with 10,000 samples. (2) the CIFAR-100 dataset \cite{krizhevsky2009learning}, containing colored images categorized into 100 classes such as "Apple" and "Dolphin". Each class has exactly 500 training samples and 100 test samples, making a total of 50,000 training samples and 10,000 test samples for performance evaluation.        
    \item{FL configuration:} For both the CIFAR-10 and CIFAR-100 datasets, we initially distribute the training data to 100 clients using a Dirichlet distribution. Notably, the data distribution at the clients is non-IID, which is commonly observed in real-world systems. We model this data heterogeneity using a Dirichlet distribution with a concentration parameter $\alpha$, where lower values of $\alpha$ result in more heterogeneous data partitions. In each training round, we randomly select 10 clients to participate. We use ResNet-18 as the AI model for training, with a batch size of 64 and the SGD optimizer for local training. Each client performs 5 local epochs, and the learning rate is set to 1e-4. 
        \item{Generation Settings:} We download the "CompVis/stablediffusion-v1-5" pre-trained Stable Diffusion model checkpoint to generate the images. We adopt 20 inference steps and a guidance scale of 7. For the generated CIFAR-10 dataset, data is produced at a rate of 10 samples per round, with a maximum limit of 300 samples. For the generated CIFAR-100 dataset, the generation rate is 3 samples per round, with a maximum of 100 samples. 
    
        \item{Prompt Templates:} Data diversity has been shown to enhance model generalization in both visual and language domains\cite{morafah2024stable}. Employing diverse prompts is a method to generate a wider variety of images. In our experiment, we utilize a diverse prompt design, where for each image generation, a prompt template is randomly selected from a pool of options.
    \end{itemize}
    \subsubsection{Performance Analysis}
        As shown in Fig. \ref{fig:cifar10-acc} and Fig. \ref{fig:cifar100-acc}, the proposed GenFL method consistently achieves the highest accuracy across various settings. The model trained exclusively on AIGC-only datasets performs poorly, primarily due to the differences between AIGC-generated images and real-world images. In contrast, FL based solely on local data does not yield the best results either, mainly due to data heterogeneity. As the parameter $\alpha$ increases and data heterogeneity decreases, the performance of FL steadily improves. \par
        The comparison results show that the GenFL method achieves faster convergence at smaller $\alpha$ values; however, this acceleration diminishes as data heterogeneity decreases. We observe that while AIGC-generated data can accelerate model convergence, there is a bottleneck in the achievable accuracy, suggesting significant room for improvement in the quality of AIGC-generated images. Although FL typically converges more slowly when training exclusively on local data, the quality of local data is generally superior to that of purely AIGC-generated data. Consequently, FL based on local data can ultimately reach higher accuracy than training with only AIGC data.\par
        
        In our experiments with CIFAR-100, we observe that the AIGC-only approach converges significantly faster than traditional FL, particularly under various local data heterogeneity distributions, compared to the one with CIFAR-10. This suggests that our proposed method is more effective in scenarios with a larger number of classes. When the number of classes is small, it is easier to collect data that covers all labels during each training round. However, when the number of classes is large, the same number of participating users may not be able to provide data that covers all labels for training.\par
        In addition, we observe that the AIGC-only approach with CIFAR-100 converges significantly faster than traditional FL, particularly under various local data heterogeneity distributions, compared to the one with CIFAR-10. This suggests that our proposed method is more effective in scenarios with a larger number of classes. One possible explanation is that, when the number of classes is small, the FL aggregated model achieves more balanced learning across each class, with large and high-quality datasets that can quickly capture the features of all classes. Although AIGC generates balanced data that also promotes uniform learning progress, its relatively lower data quality means that the convergence speed of AIGC-only is not significantly faster than FL. However, when the number of classes increases, the FL aggregated model may struggle with balanced learning across all classes, potentially causing slower progress in some categories. In this case, the AIGC-only method can accelerate model convergence by generating more balanced synthetic data. Although the final accuracy of AIGC-only may not reach that of FL, its faster convergence rate can effectively speed up FL's learning process, thereby accelerating the overall model optimization and convergence.


\section{Open Research Directions}
    AIGC-assisted FL has attracted considerable attention, yet research in this field is still in its early stages. Several key research directions remain to be explored.
    
    \subsubsection{Data Generation Improvement}
    The quality of generated data is a critical factor influencing the performance of models in AIGC-assisted FL systems. To improve this quality, two potential strategies can be explored: prompt engineering and AIGC model fine-tuning.\par
    Effective prompt engineering can help reduce ambiguity in generated content. For example, a prompt like "apple" could generate images of either the fruit or "Apple" Inc., while "beetle" might result in images of either beetles or the "Beetle" Volkswagen car. This ambiguity can affect the relevance and accuracy of the generated images, making it essential to refine the prompt design. By optimizing the prompts, we can significantly improve the quality of generated images, which enhances the accuracy of the AIGC-assisted model and, in turn, boosts the overall performance of the AIGC-assisted FL system.\par
    We can also consider implementing more effective data generation strategies to enhance the quality of the generated data. One approach involves fine-tuning a small set of collected real data prior to generating additional data. Another viable option is to utilize distributed models for fine-tuning. However, both methods may introduce additional overhead for the entire AIGC-assisted FL system. Thus, it is important to investigate whether these increased costs can genuinely improve the quality of the generated data and enhance the overall performance of GenFL, as well as whether such an investment is warranted. When assessing the effectiveness of these strategies, it is crucial to balance the enhancement of data quality with the associated system costs to ensure that the final optimization is rational and efficient. Furthermore, we should take into account the specific requirements of real-world application scenarios and explore whether there are other, more cost-effective alternatives available.
        
    \subsubsection{Weighted Policy Design}
    In the context of AIGC-assisted FL, the design of weight strategies is crucial. During model aggregation, it is important to allocate weights between the FL model and the augmented model effectively. This design should be based on data quality, considering the potential differences in data from various sources, assigning higher weights to high-quality data to enhance the performance of the final model. Furthermore, the weights should be adjusted according to the contribution of each participant in the model training, particularly when a client provides data that is highly representative or accurate, in which case its weight may be increased. In addition, after each round of aggregation, the performance of FL model and the augmented model can be evaluated, and their weights are dynamically adjusted based on performance metrics to optimize subsequent aggregation results. When designing the weight strategy, it is also necessary to consider the computational resources required by each model, ensuring that resources are allocated reasonably while maintaining model performance. By combining these strategies, it is possible to effectively optimize the aggregation results of the FL model and the augmented model, thereby enhancing the overall system performance.
        
    \subsubsection{Incentive Mechanism Design}
    In AIGC-assisted FL, incentive mechanism design plays a crucial role. For instance, it is essential to incentivize clients to share labels in order to collect more useful information, while also encouraging them to generate data to enhance model training effectiveness. The incentive mechanisms should establish reasonable reward criteria to motivate clients to actively contribute data. Additionally, the design must include methods for assessing and validating data quality to avoid negative impacts on model performance due to low-quality data. It is also important to consider the resource constraints of participating clients, ensuring that incentive measures do not impose excessive burdens. Through effective mechanism design, the overall efficiency and effectiveness of the AIGC-assisted federated learning system can be significantly improved.
        
    \subsubsection{Resource Allocation Strategy}
     Dynamic resource allocation is essential to optimize the distribution of bandwidth and computing resources, thereby enhancing data transmission and processing efficiency within AIGC-assisted FL system. Furthermore, integrating data quality assessment and the volume of generated data with resource allocation strategies ensures that high-quality data is prioritized during training, improving the model's accuracy and robustness. Through exploring these research directions, AIGC-assisted federated learning systems can achieve better performance in dynamic and resource-constrained environments.
    
\section{Conclusion}
    This paper presents a detailed exploration of the AIGC-assist FL system, emphasizing its role in enhancing FL through the integration of AIGC. We address the critical challenge of data heterogeneity which limits FL's performance, and demonstrate how AIGC serves as an innovative data synthesis technique to improve model effectiveness. Our analysis includes a comprehensive overview of the system architecture, performance metrics, and challenges inherent in AIGC-assisted FL. We then introduce the GenFL architecture, where AIGC services are deployed on the server side. Through experiments conducted on CIFAR10 and CIFAR100 datasets, we illustrate that GenFL achieves faster convergence and higher accuracy, effectively mitigating the bottlenecks caused by data heterogeneity. This study underscores the significant potential of GenFL in advancing FL systems and sets the stage for future research directions that could further optimize AIGC integration in various data environments.
\bibliographystyle{IEEEtran}
\bibliography{main.bib}
	
\end{document}